\PassOptionsToPackage{unicode}{hyperref}
\PassOptionsToPackage{hyphens}{url}
\documentclass[
]{article}
\usepackage{amsmath,amssymb}
\usepackage{iftex}
\ifPDFTeX
  \usepackage[T1]{fontenc}
  \usepackage[utf8]{inputenc}
  \usepackage{textcomp} % provide euro and other symbols
\else % if luatex or xetex
  \usepackage{unicode-math} % this also loads fontspec
  \defaultfontfeatures{Scale=MatchLowercase}
  \defaultfontfeatures[\rmfamily]{Ligatures=TeX,Scale=1}
\fi
\usepackage{lmodern}
\ifPDFTeX\else
\fi
\IfFileExists{upquote.sty}{\usepackage{upquote}}{}
\IfFileExists{microtype.sty}{% use microtype if available
  \usepackage[]{microtype}
  \UseMicrotypeSet[protrusion]{basicmath} % disable protrusion for tt fonts
}{}
\makeatletter
\@ifundefined{KOMAClassName}{% if non-KOMA class
  \IfFileExists{parskip.sty}{%
    \usepackage{parskip}
  }{% else
    \setlength{\parindent}{0pt}
    \setlength{\parskip}{6pt plus 2pt minus 1pt}}
}{% if KOMA class
  \KOMAoptions{parskip=half}}
\makeatother
\usepackage{xcolor}
\usepackage[margin=1in]{geometry}
\usepackage{longtable,booktabs,array}
\usepackage{calc} % for calculating minipage widths
\usepackage{etoolbox}
\makeatletter
\patchcmd\longtable{\par}{\if@noskipsec\mbox{}\fi\par}{}{}
\makeatother
\IfFileExists{footnotehyper.sty}{\usepackage{footnotehyper}}{\usepackage{footnote}}
\makesavenoteenv{longtable}
\usepackage{graphicx}
\makeatletter
\def\maxwidth{\ifdim\Gin@nat@width>\linewidth\linewidth\else\Gin@nat@width\fi}
\def\maxheight{\ifdim\Gin@nat@height>\textheight\textheight\else\Gin@nat@height\fi}
\makeatother
\setkeys{Gin}{width=\maxwidth,height=\maxheight,keepaspectratio}
\makeatletter
\def\fps@figure{htbp}
\makeatother
\providecommand{\tightlist}{%
  \setlength{\itemsep}{0pt}\setlength{\parskip}{0pt}}
\newlength{\cslhangindent}
\newlength{\csllabelwidth}
\newlength{\cslentryspacingunit} % times entry-spacing
\newenvironment{CSLReferences}[2] % #1 hanging-ident, #2 entry spacing
 {% don't indent paragraphs
  \setlength{\parindent}{0pt}
  \ifodd #1
  \let\oldpar\par
  \def\par{\hangindent=\cslhangindent\oldpar}
  \fi
  \setlength{\parskip}{#2\cslentryspacingunit}
 }%
 {}
\usepackage{calc}

\newcommand{\CSLLeftMargin}[1]{\parbox[t]{\csllabelwidth}{#1}}
\newcommand{\CSLRightInline}[1]{\parbox[t]{\linewidth - \csllabelwidth}{#1}\break}

\ifLuaTeX
  \usepackage{selnolig}  % disable illegal ligatures
\fi
\IfFileExists{bookmark.sty}{\usepackage{bookmark}}{\usepackage{hyperref}}
\IfFileExists{xurl.sty}{\usepackage{xurl}}{} % add URL line breaks if available
\hypersetup{
  pdftitle={Generalizable, Fast, and Accurate DeepQSPR with fastprop},
  hidelinks,
  pdfcreator={LaTeX via pandoc}}

\title{Generalizable, Fast, and Accurate DeepQSPR with
\texttt{fastprop}}
\usepackage{authblk}
\usepackage{orcidlink}

\author[1]{Jackson W. Burns \orcidlink{0000-0002-0657-9426}}
\author[1,*]{William H. Green \orcidlink{0000-0003-2603-9694}}

\affil[1]{Massachusetts Institute of Technology, Cambridge, MA}
\affil[*]{Corresponding: whgreen@mit.edu}
\date{2 April, 2024}

\begin{document}
\maketitle

\hypertarget{abstract}{%
\section{Abstract}\label{abstract}}

Quantitative Structure-Property Relationship studies (QSPR), often
referred to interchangeably as QSAR, seek to establish a mapping between
molecular structure and an arbitrary target property. Historically this
was done on a target-by-target basis with new descriptors being devised
to \emph{specifically} map to a given target. Today software packages
exist that calculate thousands of these descriptors, enabling general
modeling typically with classical and machine learning methods. Also
present today are learned representation methods in which deep learning
models generate a target-specific representation during training. The
former requires less training data and offers improved speed and
interpretability while the latter offers excellent generality, while the
intersection of the two remains under-explored.

This paper introduces \texttt{fastprop}, a software package and general
Deep-QSPR framework that combines a cogent set of molecular descriptors
with deep learning to achieve state-of-the-art performance on datasets
ranging from tens to tens of thousands of molecules. \texttt{fastprop}
provides both a user-friendly Command Line Interface and highly
interoperable set of Python modules for the training and deployment of
feedforward neural networks for property prediction. This approach
yields improvements in speed and interpretability over existing methods
while statistically equaling or exceeding their performance across most
of the tested benchmarks. \texttt{fastprop} is designed with Research
Software Engineering best practices and is free and open source, hosted
at github.com/jacksonburns/fastprop.

\hypertarget{scientific-contribution}{%
\subsection{Scientific Contribution}\label{scientific-contribution}}

\texttt{fastprop} is a QSPR framework that achieves state-of-the-art
accuracy on datasets of all sizes without sacrificing speed or
interpretability. As a software package \texttt{fastprop} emphasizes
Research Software Engineering best practices, reproducibility, and ease
of use for experts across domains.

\hypertarget{keywords}{%
\section{Keywords}\label{keywords}}

\begin{itemize}
\tightlist
\item
  QSPR
\item
  Learned Representations
\item
  Deep Learning
\item
  Molecular Descriptors
\end{itemize}

\hypertarget{introduction}{%
\section{Introduction}\label{introduction}}

Chemists have long sought a method to relate only the connectivity of a
molecule to its corresponding molecular properties. The Quantitative
Structure-Property Relationship (QSPR) would effectively solve the
forward problem of molecular engineering and enable rapid development.
Reviews on the topic are numerous and cover an enormous range of
scientific subdomains; a comprehensive review of the literature is
beyond the scope of this publication, though the work of Muratov and
coauthors {[}1{]} provides an excellent starting point for further
review. An abridged version of the history behind QSPR is presented here
to contextualize the approach of \texttt{fastprop}.

\hypertarget{historical-approaches}{%
\subsection{Historical Approaches}\label{historical-approaches}}

Early in the history of computing, limited computational power meant
that significant human expertise was required to guide QSPR models
toward effectiveness. This materialized in the form of bespoke molecular
descriptors - scalar-valued functions which operate on the molecular
graph in such a way to reflect relevant structural and electronic
information. Examples include rudimentary counting descriptors, the
Wiener Index in 1947 {[}2{]}, Atom-Bond Connectivity indices in 1998
{[}3{]}, and many others {[}4{]}. To this day descriptors are still
being developed - the geometric-harmonic-Zagreb degree based descriptors
were proposed by Arockiaraj et al.~in 2023 {[}5{]}. This time consuming
technique is of course highly effective but the dispersed nature of this
chemical knowledge means that these descriptors are spread out
throughout many journals and domains with no single source to compute
them all.

The range of regression techniques applied to these descriptors has also
been limited. As explained by Muratov et. al {[}1{]} QSPR uses linear
methods (some of which are now called machine learning) almost
exclusively. The over-reliance on this category of approaches may be due
to priorities; domain experts seek interpretability in their work,
especially given that the inputs are physically meaningful descriptors,
and linear methods lend themselves well to this approach. Practice may
also have been a limitation, since historically training and deploying
neural networks required more computer science expertise than linear
methods.

All of this is not to say that Deep Learning (DL) has \emph{never} been
applied to QSPR. Applications of DL to QSPR, i.e.~DeepQSPR, were
attempted throughout this time period but focused on the use of
molecular fingerprints rather than descriptors. This may be at least
partially attributed to knowledge overlap between deep learning experts
and this sub-class of descriptors. Molecular fingerprints are bit
vectors which encode the presence or absence of sub-structures in an
analogous manner to the “bag of words” featurization strategy common to
natural language processing. Experts have bridged this gap to open this
subdomain and proved its effectiveness. In Ma and coauthors’ review of
DL for QSPR {[}6{]}, for example, it is claimed that DL with fingerprint
descriptors is more effective than with molecular-level descriptors.
They also demonstrate that DL outperforms or at least matches classical
machine learning methods across a number of ADME-related datasets. The
results of this study demonstrate that molecular-level descriptors
actually \emph{are} effective and reaffirm that DL matches or
outperforms baselines, in this case linear.

Despite their differences, both classical- and Deep-QSPR shared a lack
of generalizability. Beyond the domains of chemistry where many of the
descriptors had been originally devised, models were either unsuccessful
or more likely simply never evaluated. As interest began to shift toward
the prediction of molecular properties which were themselves descriptors
(i.e.~derived from quantum mechanics simulations) - to which none of the
devised molecular descriptors were designed to be correlated - learned
representations (LRs) emerged.

\hypertarget{shift-to-learned-representations}{%
\subsection{Shift to Learned
Representations}\label{shift-to-learned-representations}}

The exact timing of the transition from fixed descriptors
(molecular-level or fingerprints) to LRs is difficult to ascertain
{[}7{]}. Among the most cited at least is the work of Yang and coworkers
in 2019 {[}8{]} which laid the groundwork for applying LRs to “Property
Prediction” - QSPR by another name. In short, the basic idea is to
initialize a molecular graph with only information about its bonds and
atoms such as order, aromaticity, atomic number, etc. Then via a Message
Passing Neural Network (MPNN) architecture, which is able to aggregate
these atom- and bond-level features into a vector in a manner which can
be updated, the ‘best’ representation of the molecule is found during
training. This method proved highly accurate \emph{and} achieved the
generalizability apparently lacking in descriptor-based modeling. The
modern version of the corresponding software package Chemprop (described
in {[}9{]}) has become a \emph{de facto} standard for property
prediction, partially because of the significant development and
maintenance effort supporting that open source software project.

Following the initial success of Chemprop numerous representation
learning frameworks have been devised, all of which slightly improve
performance. The Communicative-MPNN (CMPNN) framework is a modified
version of Chemprop with a different message passing scheme to increase
the interactions between node and edge features {[}10{]}. Uni-Mol
incorporates 3D information and relies extensively on transformers
{[}11{]}. In a “full circle moment” architectures like the Molecular
Hypergraph Neural Network (MHNN) have been devised to learn
representations for specific subsets of chemistry, in that case
optoelectronic properties {[}12{]}. Myriad others exist including
GSL-MPP (accounts for intra-dataset molecular similarity) {[}13{]},
SGGRL (trains three representations simultaneously using different input
formats) {[}14{]}, and MOCO (multiple representations and contrastive
pretraining) {[}15{]}.

\hypertarget{limitations}{%
\subsubsection{Limitations}\label{limitations}}

Despite the continuous incremental performance improvements, this area
of research has had serious drawbacks. A thru-theme in these frameworks
is the increasing complexity of DL techniques and consequent
un-interpretability. This also means that actually \emph{using} these
methods to do research on real-world dataset requires varying amounts of
DL expertise, creating a rift between domain experts and these methods.
Perhaps the most significant failure is the inability to achieve good
predictions on small \footnote{What constitutes a ‘small’ dataset is
  decidedly \emph{not} agreed upon by researchers. For the purposes of
  this study, it will be used to refer to datasets with
  \textasciitilde1000 molecules or fewer, which the authors believe
  better reflects the size of real-world datasets.} datasets. This is a
long-standing limitation, with the original Chemprop paper stating that
linear models are about on par with Chemprop for datasets with fewer
than 1000 entries {[}8{]}.

This limitation is especially challenging because it is a
\emph{fundamental} drawback of the LR approach. Without the use of
advanced DL techniques like pre-training or transfer learning, the model
is essentially starting from near-zero information every time a model is
created. This inherently requires larger datasets to allow the model to
effectively ‘re-learn’ the chemical intuition which was built in to
descriptor- and fixed fingerprint-based representations.

Efforts are of course underway to address this limitation, though none
are broadly successful. One simple but incredibly computationally
expensive approach is to use delta learning, which artificially
increases dataset size by generating all possible \emph{pairs} of
molecules from the available data (thus squaring the size of the
dataset). This was attempted by Nalini et al. {[}16{]}, who used an
unmodified version of Chemprop referred to as ‘DeepDelta’ to predict
\emph{differences} in molecular properties for \emph{pairs} of
molecules. They achieve increased performance over standard LR
approaches but \emph{lost} the ability to train on large datasets due to
simple runtime limitations. Another promising line of inquiry is the
Transformer-CNN model of Karpov et al. {[}17{]}, which leverages a
pre-trained transformer model for prediction, circumventing the need for
massive datasets and offering additional benefits in interpretability.
This model is unique in that it operates directly on the SMILES
representation of the molecule, also offering benefits in structural
attribution of predictions. Other increasingly complex approaches are
discussed in the outstanding review by van Tilborg et al. {[}18{]}.

While iterations on LRs and novel approaches to low-data regimes have
been in development, the classical QSPR community has continued their
work. A turning point in this domain was the release of
\texttt{mordred}, a fast and well-developed package capable of
calculating more than 1600 molecular descriptors {[}19{]}. Critically
this package was fully open source and written in Python, allowing it to
readily interoperate with the world-class Python DL software ecosystem
that greatly benefitted the LR community. Despite previous claims that
molecular descriptors \emph{cannot} achieve generalizable QSPR in
combination with DL, the opposite is shown here with \texttt{fastprop}.

\hypertarget{implementation}{%
\section{Implementation}\label{implementation}}

\begin{figure}
\centering
\includegraphics[width=2in,height=\textheight]{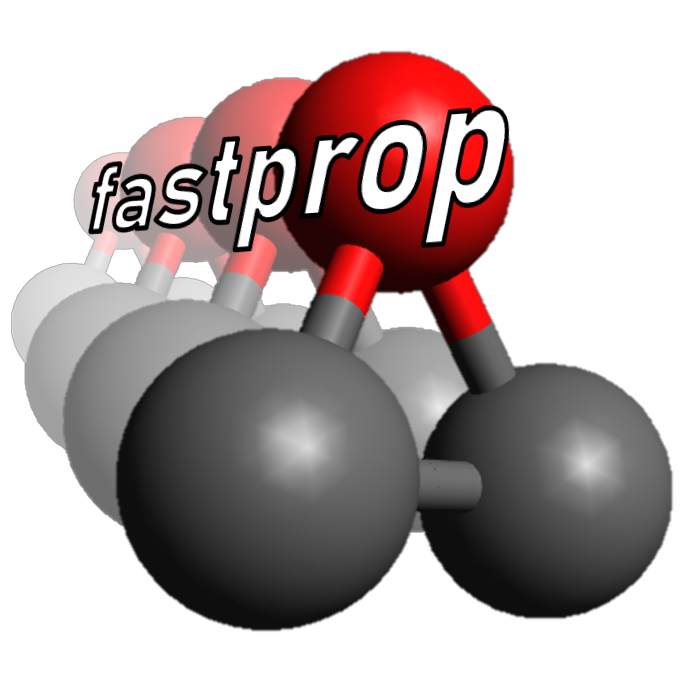}
\caption{\texttt{fastprop} logo.\label{logo}}
\end{figure}

At its core the \texttt{fastprop} ‘architecture’ is simply the
\texttt{mordred} molecular descriptor calculator \footnote{The original
  \texttt{mordred} package is no longer maintained. \texttt{fastprop}
  uses a fork of \texttt{mordred} called \texttt{mordredcommunity} that
  is maintained by community-contributed patches (see
  github.com/JacksonBurns/mordred-community).} {[}19{]} connected to a
Feedforward Neural Network (FNN) implemented in PyTorch Lightning
{[}20{]} (Figure \ref{logo}) - an existing approach formalized into an
easy-to-use, reliable, and correct implementation. \texttt{fastprop} is
highly modular for seamless integration into existing workflows and
includes and end-to-end Command Line Interface (CLI) for general use. In
the latter mode the user simply specifies a set of SMILES {[}21{]}, a
linear textual encoding of molecules, and their corresponding target
values. \texttt{fastprop} optionally standardizes input molecule and
then automatically calculates and caches the corresponding molecular
descriptors with \texttt{mordred}, re-scales both the descriptors and
the targets appropriately, and then trains an FNN to predict the
indicated targets. By default this FNN is two hidden layers with 1800
neurons each connected by ReLU activation functions, though the
configuration can be readily changed via the CLI or configuration file.
Multitask regression and multi-label classification are also supported
and configurable in the same manner, the former having been shown to
significantly improve predictive power in cheminformatics models
{[}22{]}. \texttt{fastprop} principally owes its success to the cogent
set of descriptors assembled by the developers of \texttt{mordred}.
Multiple descriptor calculators from the very thorough review by
McGibbon et al. {[}23{]} could be used instead, though none are as
readily interoperable as \texttt{mordred}. Additionally, the ease of
training FNNs with modern software like PyTorch Lightning and the
careful application of Research Software Engineering best practices make
\texttt{fastprop} as user friendly as the best-maintained alternatives.

This trivially simple idea has been alluded to in previous published
work but neither described in detail nor lauded for its generalizability
or accuracy. Comesana and coauthors, based on a review of the biofuels
property prediction landscape, claimed that methods (DL or otherwise)
using large numbers of molecular descriptors were unsuccessful, instead
proposing a feature selection method {[}24{]}. As a baseline in a study
of photovoltaic property prediction, Wu et al.~reported using the
\texttt{mordred} descriptors in combination with both a Random Forest
and an Artificial Neural Network, though in their hands the performance
is worse than their bespoke model and no code is available for
inspection {[}25{]}.

Others have also incorporated \texttt{mordred} descriptors into their
modeling efforts, though none with a simple FNN as described above.
Esaki and coauthors started a QSPR study with \texttt{mordred}
descriptors for a dataset of small molecules, but via an enormously
complex modeling pipeline (using only linear methods) removed all but 53
{[}26{]}. Yalamanchi and coauthors used DL on \texttt{mordred}
descriptors as part of a two-headed representation, but their network
architecture was sequential hidden layers \emph{decreasing} in size to
only 12 features {[}27{]} as opposed to the constant 1800 in
\texttt{fastprop}.

The reason \texttt{fastprop} stands out from these studies and
contradicts previous reports is for the simple reason that it works. As
discussed at length in the
\protect\hyperlink{results--discussion}{Results \& Discussion} section,
this approach statistically matches or exceeds the performance of
leading LR approaches on common benchmark datasets and bespoke QSPR
models on small real-world datasets. \texttt{fastprop} also overcomes
the limitations of LRs discussed above. The FNN architecture and use of
physically meaningful molecular descriptors enables the application of
SHAP {[}28{]}, a common tool for feature importance analysis (see
\protect\hyperlink{interpretability}{Interpretability}). The simplicity
of the framework enables domain experts to apply it easily and makes
model training dramatically faster than LRs. Most importantly this
approach is successful on the \emph{smallest} of real-world datasets. By
starting from such an informed initialization the FNN can be readily
trained on datasets with as few as \emph{forty} training examples (see
\protect\hyperlink{pahs}{PAHs}).

\hypertarget{example-usage}{%
\subsection{Example Usage}\label{example-usage}}

\texttt{fastprop} is built with ease of use at the forefront of design.
To that end, input data is accepted in the immensely popular
Comma-Separated Values (CSV) format, editable with all modern
spreadsheet editors and completely platform independent. An example
specify some properties for benzene is shown below, including its SMILES
string:

\begin{verbatim}
compound,smiles,log_p,retention_index,boiling_point_c,acentric_factor
Benzene,C1=CC=CC=C1,2.04,979,80,0.21
\end{verbatim}

\texttt{fastprop} itself is accessed via the command line interface,
with configuration parameters passed as either command line arguments or
in an easily editable configuration file:

\begin{verbatim}
# pah.yml
# generic args
output_directory: pah
random_seed: 55
problem_type: regression
# featurization
input_file: pah/arockiaraj_pah_data.csv
target_columns: log_p
smiles_column: smiles
descriptor_set: all
# preprocessing
clamp_input: True
# training
number_repeats: 4
number_epochs: 100
batch_size: 64
patience: 15
train_size: 0.8
val_size: 0.1
test_size: 0.1
sampler: random
\end{verbatim}

Training, prediction, and feature importance are then readily accessible
via the commands \texttt{fastprop\ train}, \texttt{fastprop\ predict},
or \texttt{fastprop\ shap}, respectively. The \texttt{fastprop} GitHub
repository contains a Jupyter notebook runnable from the browser via
Google colab which allows users to actually execute the above example,
which is also discussed at length in the \protect\hyperlink{pah}{PAHs
section}, as well as further details about each configurable option.

\hypertarget{results-discussion}{%
\section{Results \& Discussion}\label{results-discussion}}

There are a number of established molecular property prediction
benchmarks commonly used in LR studies, especially those standardized by
MoleculeNet {[}29{]}. Principal among them are QM8 {[}30{]} and QM9
{[}31{]}, often regarded as \emph{the} standard benchmark for property
prediction. These are important benchmarks and QM9 is included for
completeness, though the enormous size and rich coverage of chemical
space in the QM9 dataset means that nearly all model architectures are
highly accurate, including \texttt{fastprop}.

Real world experimental datasets, particularly those common in QSPR
studies, often number in the hundreds. To demonstrate the applicability
of \texttt{fastprop} to these regimes, many smaller datasets are
selected including some from the QSPR literature that are not
established benchmarks. These studies relied on more complex and slow
modeling techniques (\protect\hyperlink{ara}{ARA}) or the design of a
bespoke descriptor (\protect\hyperlink{pah}{PAHs}) and have not yet come
to rely on learned representations as a go-to tool. In these
data-limited regimes where LRs sometimes struggle, \texttt{fastprop} and
its intuition-loaded initialization are highly powerful. To emphasize
this point further, the benchmarks are presented in order of dataset
size, descending.

Two additional benchmarks showing the limitations of \texttt{fastprop}
are included after the main group of benchmarks: Fubrain and
QuantumScents. The former demonstrates how \texttt{fastprop} can
outperform LRs but still trail approaches like delta learning. The later
is a negative result showing how \texttt{fastprop} can fail on
especially difficult, atypical targets.

All of these \texttt{fastprop} benchmarks are reproducible, and complete
instructions for installation, data retrieval and preparation, and
training are publicly available on the \texttt{fastprop} GitHub
repository at
\href{https://github.com/jacksonburns/fastprop}{github.com/jacksonburns/fastprop}.

\hypertarget{benchmark-methods}{%
\subsection{Benchmark Methods}\label{benchmark-methods}}

The method for splitting data into training, validation, and testing
sets varies on a per-study basis and is described in each sub-section.
Sampling is performed using the \texttt{astartes} package {[}32{]} which
implements a variety of sampling algorithms and is highly reproducible.
For datasets containing missing target values or invalid SMILES strings,
those entries were dropped, as is the default behavior of
\texttt{fastprop}.

Results for \texttt{fastprop} are reported as the average value of a
metric and its standard deviation across a number of repetitions
(repeated re-sampling of the dataset). The number of repetitions is
chosen to either match referenced literature studies or else increased
from two until the performance no longer meaningfully changes. Note that
this is \emph{not} the same as cross-validation. Each section also
includes the performance of a zero-layer (i.e.~linear regression)
network as a baseline to demonstrate the importance of non-linearity in
a deep NN.

For performance metrics retrieved from literature it is assumed that the
authors optimized their respective models to achieve the best possible
results; therefore, \texttt{fastprop} metrics are reported after model
optimization using the \texttt{fastprop\ train\ ...\ -\/-optimize}
option. When results are generated for this study using Chemprop, the
default settings are used except that the number of epochs is increased
to allow the model to converge and batch size is increased to match
dataset size and speed up training. Chemprop was chosen as a point of
comparison throughout this study since it is among the most accessible
and well-maintained software packages for molecular machine learning.

When reported, execution time is as given by the unix \texttt{time}
command using Chemprop version 1.6.1 on Python 3.8 and includes the
complete invocation of Chemprop,
i.e.~\texttt{time\ chemprop\_train\ ...}. The insignificant time spent
manually collating Chemprop results (Chemprop does not natively support
repetitions) is excluded. \texttt{fastprop} is run on version 1.0.6
using Python 3.11 and timing values are reported according to its
internal time measurement which was verified to be nearly identical to
the Unix \texttt{time} command. The coarse comparison of the two
packages is intended to emphasize the scaling of LRs and Deep-QSPR and
that \texttt{fastprop} is, generally speaking, much faster. All models
trained for this study were run on a Dell Precision series laptop with
an NVIDIA Quadro RTX 4000 GPU and Intel Xeon E-2286M CPU.

Because the diversity of methods across these different datasets
complicates inter-dataset comparisons, an additional set of benchmarks
using an identical method across all datasets is included in Additional
File 1, Table S1. This benchmark compares \texttt{fastprop}, Chemprop,
and the aforementioned Transformer-CNN, which is especially suitable for
the small datasets included therein. No long-form commentary on
benchmark results is provided, though the conclusions are largely the
same as those shown here.

\hypertarget{performance-metrics}{%
\subsubsection{Performance Metrics}\label{performance-metrics}}

The evaluation metrics used in each of these benchmarks are chosen to
match literature precedent, particularly as established by MoleculeNet
{[}29{]}, where available. It is common to use scale-dependent metrics
that require readers to understand the relative magnitude of the target
variables. The authors prefer more readily interpretable metrics such as
(Weighted) Mean Absolute Percentage Error (W/MAPE) and are included
where relevant.

All metrics are defined according to their typical formulae which are
readily available online and are implemented in common software
packages. Those presented here are summarized below, first for
regression:

\begin{itemize}
\tightlist
\item
  Mean Absolute Error (MAE): Absolute difference between predictions and
  ground truth averaged across dataset; scale-dependent.
\item
  Root Mean Squared Error (RMSE): Absolute differences \emph{squared}
  and then averaged; scale-dependent.
\item
  Mean Absolute Percentage Error (MAPE): MAE except that differences are
  relative (i.e.~divided by the ground truth); scale-independent, range
  0 (best) and up.
\item
  Weighted Mean Absolute Percentage Error (WMAPE): MAPE except the
  average is a weighted average, where the weight is the magnitude of
  the ground truth; scale-independent, range 0 (best) and up.
\item
  Coefficient of Determination (R2): Proportion of variance explained;
  scale-independent, range 0 (worst) to 1 (best).
\end{itemize}

and classification: - Area Under the Receiver Operating Curve (AUROC,
AUC, or ROC-AUC): Summary statistic combining all possible
classification errors; scale-independent, range 0.5 (worst, random
guessing) to 1.0 (perfect classifier). - Accuracy: Fraction of correct
classifications, expressed as a percentage; scale-independent, range 0
(worst) to 100 (perfect classifier).

\hypertarget{benchmark-results}{%
\subsection{Benchmark Results}\label{benchmark-results}}

See Table \ref{results_table} for a summary of all the results.
Subsequent sections explore each in greater detail. For benchmarks with
statistics, practically significant best performers are shown in bold.

\begin{longtable}[]{@{}
  >{\raggedright\arraybackslash}p{(\columnwidth - 12\tabcolsep) * \real{0.1368}}
  >{\raggedright\arraybackslash}p{(\columnwidth - 12\tabcolsep) * \real{0.1795}}
  >{\raggedright\arraybackslash}p{(\columnwidth - 12\tabcolsep) * \real{0.1197}}
  >{\raggedright\arraybackslash}p{(\columnwidth - 12\tabcolsep) * \real{0.1282}}
  >{\raggedright\arraybackslash}p{(\columnwidth - 12\tabcolsep) * \real{0.1111}}
  >{\raggedright\arraybackslash}p{(\columnwidth - 12\tabcolsep) * \real{0.2222}}
  >{\raggedright\arraybackslash}p{(\columnwidth - 12\tabcolsep) * \real{0.0598}}@{}}
\caption{Summary of benchmark results, best state-of-the-art method
vs.~\texttt{fastprop} and Chemprop.
\label{results_table}}\tabularnewline
\toprule\noalign{}
\begin{minipage}[b]{\linewidth}\raggedright
Benchmark
\end{minipage} & \begin{minipage}[b]{\linewidth}\raggedright
Samples
\end{minipage} & \begin{minipage}[b]{\linewidth}\raggedright
Metric
\end{minipage} & \begin{minipage}[b]{\linewidth}\raggedright
\begin{verbatim}
SOTA
\end{verbatim}
\end{minipage} & \begin{minipage}[b]{\linewidth}\raggedright
\texttt{fastprop}
\end{minipage} & \begin{minipage}[b]{\linewidth}\raggedright
\begin{verbatim}
   Chemprop
\end{verbatim}
\end{minipage} & \begin{minipage}[b]{\linewidth}\raggedright
p
\end{minipage} \\
\midrule\noalign{}
\endfirsthead
\toprule\noalign{}
\begin{minipage}[b]{\linewidth}\raggedright
Benchmark
\end{minipage} & \begin{minipage}[b]{\linewidth}\raggedright
Samples
\end{minipage} & \begin{minipage}[b]{\linewidth}\raggedright
Metric
\end{minipage} & \begin{minipage}[b]{\linewidth}\raggedright
\begin{verbatim}
SOTA
\end{verbatim}
\end{minipage} & \begin{minipage}[b]{\linewidth}\raggedright
\texttt{fastprop}
\end{minipage} & \begin{minipage}[b]{\linewidth}\raggedright
\begin{verbatim}
   Chemprop
\end{verbatim}
\end{minipage} & \begin{minipage}[b]{\linewidth}\raggedright
p
\end{minipage} \\
\midrule\noalign{}
\endhead
\bottomrule\noalign{}
\endlastfoot
QM9 & 133,885 & MAE & 0.0047\(^a\) & 0.0060 & 0.0081\(^a\) &
\textasciitilde{} \\
Pgp & 1,275 & AUROC & 0.94\(^b\) & 0.90 & 0.89\(^b\) &
\textasciitilde{} \\
ARA & 842 & Accuracy & 91\(^c\) & 88 & 82* & 0.083 \\
Flash & 632 & RMSE & 13.2\(^d\) & \textbf{13.0} & 21.2* & 0.021 \\
YSI & 442 & MAE & 22.3\(^e\) & 25.0 & 28.9* & 0.29 \\
PAH & 55 & R2 & 0.96\(^f\) & \textbf{0.97} & 0.59* & 0.0012 \\
\end{longtable}

a {[}11{]} b {[}33{]} c {[}34{]} d {[}35{]} e {[}36{]} f {[}5{]} * These
reference results were generated for this study.

Statistical comparisons of \texttt{fastprop} to Chemprop (shown in the
\texttt{p} column) are performed using the non-parametric
Wilcoxon-Mann-Whitney Test as implemented in GNumeric. Values are only
shown for results generated in this study which are known to be
performed using the same methods. Only the results for Flash and PAH are
statistically significant at 95\% confidence (p\textless0.05), see
benchmark-specific subsections for confidence intervals.

\hypertarget{qm9}{%
\subsubsection{QM9}\label{qm9}}

Originally described in Scientific Data {[}31{]} and perhaps the most
established property prediction benchmark, Quantum Machine 9 (QM9)
provides quantum mechanics derived descriptors for many small molecules
containing one to nine heavy atoms, totaling 133,885. The data was
retrieved from MoleculeNet {[}29{]} in a readily usable format. As a
point of comparison, performance metrics are retrieved from the paper
presenting the UniMol architecture {[}11{]} previously mentioned. In
that study they trained on only three especially difficult targets
(homo, lumo, and gap) using scaffold-based splitting (a more challenging
alternative to random splitting), reporting mean and standard deviation
across 3 repetitions.

\texttt{fastprop} achieves 0.0060 \(\pm\) 0.0002 mean absolute error,
whereas Chemprop achieves 0.00814 \(\pm\) 0.00001 and the UniMol
framework manages 0.00467 \(\pm\) 0.00004. This places the
\texttt{fastprop} framework ahead of previous learned representation
approaches but still trailing UniMol. This is not completely unexpected
since UniMol encodes 3D information from the dataset whereas Chemprop
and \texttt{fastprop} use only 2D. Future work could evaluate the use of
3D-based descriptors to improve \texttt{fastprop} performance in the
same manner that UniMol has with LRs. All methods are better than a
purely linear model trained on the molecular descriptors, which manages
only a 0.0095 \(\pm\) 0.0006 MAE.

\hypertarget{pgp}{%
\subsubsection{Pgp}\label{pgp}}

First reported in 2011 by Broccatelli and coworkers {[}37{]}, this
dataset has since become a standard benchmark and is included in the
Therapeutic Data Commons (TDC) {[}38{]} model benchmarking suite. The
dataset maps 1,275 small molecule drugs to a binary label indicating if
they inhibit P-glycoprotein (Pgp). TDC serves this data through a Python
package, but due to installation issues the data was retrieved from the
original study instead. The recommended splitting approach is a 70/10/20
scaffold-based split which is done here with 4 replicates.

The model in the original study uses a molecular interaction field but
has since been surpassed by other models. According to TDC the current
leader {[}33{]} on this benchmark has achieved an AUROC of 0.938 \(\pm\)
0.002 \footnote{See
  \href{https://tdcommons.ai/benchmark/admet_group/03pgp/}{the TDC Pgp
  leaderboard}.}. On the same leaderboard Chemprop {[}8{]} achieves
0.886 \(\pm\) 0.016 with the inclusion of additional molecular features.
\texttt{fastprop} yet again approaches the performance of the leading
methods and outperforms Chemprop, here with an AUROC of 0.903 \(\pm\)
0.033 and an accuracy of 83.6 \(\pm\) 4.6\%. Remarkably, the linear QSPR
model outperforms both Chemprop and \texttt{fastprop}, approaching the
performance of the current leader with an AUROC of 0.917 \(\pm\) 0.016
and an accuracy of 83.8 \(\pm\) 3.9\%.

\hypertarget{ara}{%
\subsubsection{ARA}\label{ara}}

Compiled by Schaduangrat et al.~in 2023 {[}34{]}, this dataset maps 842
small molecules to a binary label indicating if the molecule is an
Androgen Receptor Antagonist (ARA). The reference study introduced
DeepAR, a highly complex modeling approach, which achieved an accuracy
of 91.1\% and an AUROC of 0.945.

For this study an 80/10/10 random splitting is repeated four times on
the dataset since no analogous split to the reference study can be
determined. Chemprop takes 16 minutes and 55 seconds to run on this
dataset and achieves only 82.4 \(\pm\) 2.0\% accuracy and 0.898 \(\pm\)
0.022 AUROC. \texttt{fastprop} takes only 1 minute and 54 seconds (1
minute and 39 seconds for descriptor calculation) and is competitive
with the reference study in performance, achieving a 88.2 \(\pm\) 3.7\%
accuracy and 0.935 \(\pm\) 0.034 AUROC. The purely linear QSPR model
falls far behind these methods with a 71.8 \(\pm\) 6.6\% accuracy and
0.824 \(\pm\) 0.052 AUROC.

\hypertarget{flash}{%
\subsubsection{Flash}\label{flash}}

First assembled and fitted to by Saldana and coauthors {[}35{]} the
dataset (Flash) includes around 632 entries, primarily alkanes and some
oxygen-containing compounds, and their literature-reported flash point.
The reference study reports the performance on only one repetition, but
manually confirms that the distribution of points in the three splits
follows the parent dataset. The split itself was a 70/20/10 random
split, which is repeated four times for this study.

Using a complex multi-model ensembling method, the reference study
achieved an RMSE of 13.2, an MAE of 8.4, and an MAPE of 2.5\%.
\texttt{fastprop} matches this performance, achieving 13.0 \(\pm\) 2.0
RMSE, 9.0 \(\pm\) 0.5 MAE, and 2.7\% \(\pm\) 0.1\% MAPE. Chemprop,
however, struggles to match the accuracy of either method - it manages
an RMSE of 21.2 \(\pm\) 2.2, an MAE of 13.8 \(\pm\) 2.1, and a MAPE of
3.99 \(\pm\) 0.36\%. This is worse than the performance of the linear
QSPR model, with an RMSE of 16.1 \(\pm\) 4.0, an MAE of 11.3 \(\pm\)
2.9, and an MAPE of 3.36 \(\pm\) 0.77\%.

\texttt{fastprop} dramatically outperforms both methods in terms of
training time. The reference model required significant manual
intervention to create a model ensemble, so no single training time can
be fairly identified. \texttt{fastprop} arrived at the indicated
performance without any manual intervention in only 30 seconds, 13 of
which were spent calculating descriptors. Chemprop, in addition to not
reaching the same level of accuracy, took 5 minutes and 44 seconds to do
so - more than ten times the execution time of \texttt{fastprop}.

\hypertarget{ysi}{%
\subsubsection{YSI}\label{ysi}}

Assembled by Das and coauthors {[}36{]} from a collection of other
smaller datasets, this dataset maps 442 molecular structures to a
unified-scale Yield Sooting Index (YSI), a molecular property of
interest to the combustion community. The reference study performs
leave-one-out cross validation to fit a per-fragment contribution model,
effectively a training size of \textgreater99\%, without a holdout set.
Though this is not standard practice and can lead to overly optimistic
reported performance, the results will be carried forward regardless.
The original study did not report overall performance metrics, so they
have been re-calculated for this study using the predictions made by the
reference model as provided on GitHub \footnote{Predictions are
  available at this
  \href{https://github.com/pstjohn/ysi-fragment-prediction/blob/bdf8b16a792a69c3e3e63e64fba6f1d190746abe/data/ysi_predictions.csv}{permalink}
  to the CSV file on GitHub.}. For comparison \texttt{fastprop} and
Chemprop use a more typical 60/20/20 random split and 8 repetitions.
Results are summarized in Table \ref{ysi_results_table}.

\begin{longtable}[]{@{}
  >{\raggedright\arraybackslash}p{(\columnwidth - 6\tabcolsep) * \real{0.1806}}
  >{\raggedright\arraybackslash}p{(\columnwidth - 6\tabcolsep) * \real{0.2361}}
  >{\raggedright\arraybackslash}p{(\columnwidth - 6\tabcolsep) * \real{0.2361}}
  >{\raggedright\arraybackslash}p{(\columnwidth - 6\tabcolsep) * \real{0.2500}}@{}}
\caption{Accuracy of YSI predictions from Reference model {[}36{]},
Linear QSPR model, \texttt{fastprop}, and Chemprop.
\label{ysi_results_table}}\tabularnewline
\toprule\noalign{}
\begin{minipage}[b]{\linewidth}\raggedright
Model
\end{minipage} & \begin{minipage}[b]{\linewidth}\raggedright
MAE
\end{minipage} & \begin{minipage}[b]{\linewidth}\raggedright
RMSE
\end{minipage} & \begin{minipage}[b]{\linewidth}\raggedright
WMAPE
\end{minipage} \\
\midrule\noalign{}
\endfirsthead
\toprule\noalign{}
\begin{minipage}[b]{\linewidth}\raggedright
Model
\end{minipage} & \begin{minipage}[b]{\linewidth}\raggedright
MAE
\end{minipage} & \begin{minipage}[b]{\linewidth}\raggedright
RMSE
\end{minipage} & \begin{minipage}[b]{\linewidth}\raggedright
WMAPE
\end{minipage} \\
\midrule\noalign{}
\endhead
\bottomrule\noalign{}
\endlastfoot
Reference & 22.3 & 50 & 14 \\
\texttt{fastprop} & 25.0 \(\pm\) 5.2 & 52 \(\pm\) 20 & 13.6 \(\pm\)
1.3 \\
Chemprop & 28.9 \(\pm\) 6.5 & 63 \(\pm\) 14 & 16.4 \(\pm\) 3.0 \\
Linear & 82 \(\pm\) 39 & 180 \(\pm\) 120 & 47.0 \(\pm\) 2.3 \\
\end{longtable}

\texttt{fastprop} again outperforms Chemprop, in this case approaching
the overly-optimistic performance of the reference model. Taking into
account that reference model has been trained on a significantly larger
amount of data, this performance is admirable. Also notable is the
difference in training times. Chemprop takes 7 minutes and 2 seconds
while \texttt{fastprop} completes in only 42 seconds, again a factor of
ten faster. The linear QSPR model fails entirely, performing
dramatically worse than all other models.

\hypertarget{pah}{%
\subsubsection{PAH}\label{pah}}

Originally compiled by Arockiaraj et al. {[}5{]} the Polycyclic Aromatic
Hydrocarbons (PAH) dataset contains water/octanol partition coefficients
(logP) for 55 polycyclic aromatic hydrocarbons ranging in size from
napthalene to circumcoronene. This size of this benchmark is an ideal
case study for the application of \texttt{fastprop}. Using expert
insight the reference study designed a novel set of molecular
descriptors that show a strong correlation to logP, with correlation
coefficients ranging from 0.96 to 0.99 among the various new
descriptors.

For comparison, \texttt{fastprop} and Chemprop are trained using 8
repetitions of a typical 80/10/10 random split - only \textbf{44}
molecules in the training data. \texttt{fastprop} matches the
performance of the bespoke descriptors with a correlation coefficient of
0.972 \(\pm\) 0.025. This corresponds to an MAE of 0.19 \(\pm\) 0.10 and
an MAPE of 2.5 \(\pm\) 1.5\%. Chemprop effectively fails on this
dataset, achieving a correlation coefficient of only 0.59 \(\pm\) 0.24,
an MAE of 1.04 \(\pm\) 0.33 (one anti-correlated outlier replicate
removed). This is worse even than the purely linear QSPR model, which
manages a correlation coefficient of 0.78 \(\pm\) 0.22, an MAE of 0.59
\(\pm\) 0.22, and an RMSE of 0.75 \(\pm\) 0.32. Despite the large
parameter size of the \texttt{fastprop} model relative to the training
data, it readily outperforms Chemprop in the small-data limit.

For this unique dataset, execution time trends are inverted.
\texttt{fastprop} takes 1 minute and 43 seconds, of which 1 minute and
31 seconds were spent calculating descriptors for these unusually large
molecules. Chemprop completes in 1 minute and 16 seconds, faster overall
but much slower compared with the training time of \texttt{fastprop}
without descriptor calculation.

\hypertarget{limitations-and-future-work}{%
\section{Limitations and Future
Work}\label{limitations-and-future-work}}

\hypertarget{negative-results}{%
\subsection{Negative Results}\label{negative-results}}

The \texttt{fastprop} framework is not without its drawbacks. The two
subsequent sections explore in greater detail two specific cases where
\texttt{fastprop} loses out to existing methods, but some general notes
about out-of-distribution predictions and overfitting are also included
here. Like all machine learning methods, \texttt{fastprop} is not
intended to make predictions outside of its training feature space. The
use of molecular descriptors, which can become out-of-distribution, may
exacerbate this problem but \texttt{fastprop} can optionally winsorize
the descriptors to counteract this issue. Additionally, hyperparameter
optimization of machine learning models in cheminformatics has been
known to cause overfitting {[}39{]}, especially on small datasets. Users
should be cautious when optimizing \texttt{fastprop} models and rely on
defaults when possible.

\hypertarget{delta-learning-with-fubrain}{%
\subsubsection{Delta Learning with
Fubrain}\label{delta-learning-with-fubrain}}

First described by Esaki and coauthors, the Fraction of Unbound Drug in
the Brain (Fubrain) dataset is a collection of about 0.3k small molecule
drugs and their corresponding experimentally measured unbound fraction
in the brain, a critical metric for drug development {[}26{]}. This
specific target in combination with the small dataset size makes this
benchmark highly relevant for typical QSPR studies, particular via delta
learning. DeepDelta {[}16{]} performed a 90/0/10 cross-validation study
of the Fubrain dataset in which the training and testing molecules were
intra-combined to generate all possible pairs and then the differences
in the property \footnote{Although the original Fubrain study reported
  untransformed fractions, the DeepDelta authors confirmed
  \href{https://github.com/RekerLab/DeepDelta/issues/2\#issuecomment-1917936697}{via
  GitHub} that DeepDelta was trained on log base-10 transformed fraction
  values, which is replicated here.} were predicted, rather than the
absolute values, increasing the amount of training data by a factor of
300.

DeepDelta reported an RMSE of 0.830 \(\pm\) 0.023 at predicting
differences, whereas a typical Chemprop model trained to directly
predict property values was only able to reach an RMSE of 0.965 \(\pm\)
0.09 when evaluated on its capacity to predict property differences.
\texttt{fastprop} is able to outperform Chemprop, though not DeepDelta,
achieving an RMSE of 0.930 \(\pm\) 0.029 when using the same splitting
procedure above. It is evident that delta learning is still a powerful
technique for regressing small datasets.

For completeness, the performance of Chemprop and \texttt{fastprop} when
directly predicting the unbound fraction are also compared to the
original study by Esaki and coauthors. Using both cross validation and
and external test sets, they had an effective
training/validation/testing split of 0.64/0.07/0.28 which will be
repeated 4 times here for comparison. They used \texttt{mordred}
descriptors in their model but as is convention they strictly applied
linear modeling methods. All told, their model achieved an RMSE of 0.53
averaged across all testing data. In only 39 seconds, of which 31 are
spent calculating descriptors, \texttt{fastprop} far exceeds the
reference model with an RMSE of 0.207 \(\pm\) 0.024. This also surpasses
Chemprop, itself outperforming the reference model with an RMSE of 0.223
\(\pm\) 0.036.

\hypertarget{fastprop-fails-on-quantumscents}{%
\subsubsection{\texorpdfstring{\texttt{fastprop} Fails on
QuantumScents}{fastprop Fails on QuantumScents}}\label{fastprop-fails-on-quantumscents}}

Compiled by Burns and Rogers {[}40{]}, this dataset contains
approximately 3.5k SMILES and 3D structures for a collection of
molecules labeled with their scents. Each molecule can have any number
of reported scents from a possible 113 different labels, making this
benchmark a a Quantitative Structure-Odor Relationship. Due to the
highly sparse nature of the scent labels a unique sampling algorithm
(Szymanski sampling {[}41{]}) was used in the reference study and the
exact splits are replicated here for a fair comparison.

In the reference study, Chemprop achieved an AUROC of 0.85 with modest
hyperparameter optimization and an improved AUROC of 0.88 by
incorporating the atomic descriptors calculated as part of
QuantumScents. \texttt{fastprop} is incapable of incorporating atomic
features, so they are not included. Using only the 2D structural
information, \texttt{fastprop} falls far behind the reference study with
an AUROC of only 0.651 \(\pm\) 0.005. Even when using the high-quality
3D structures and calculating additional descriptors (demonstrated in
the GitHub repository), the performance does not improve.

The exact reason for this failure is unknown. Possible reasons include
that the descriptors in \texttt{mordred} are simply not correlated with
this target, and thus the model struggles to make predictions. This is a
fundamental drawback of this fixed representation method - whereas a LR
could adapt to this unique target, \texttt{fastprop} fails.

\hypertarget{execution-time}{%
\subsection{Execution Time}\label{execution-time}}

\texttt{fastprop} is consistently faster to train than Chemprop when
using a GPU, helping exploit the ‘time value’ of data. Note that due to
the large size of the FNN in \texttt{fastprop} it can be slower than
small Chemprop models when training on a CPU since Chemprop uses a much
smaller FNN and associated components.

There is a clear performance improvement to be had by reducing the
number of descriptors to a subset of only the most important. Future
work can address this possibility to decrease time requirements for both
training by reducing network size and inference by decreasing the number
of descriptors to be calculated for new molecules. This has \emph{not}
been done in this study for two reasons: (1) to emphasize the capacity
of the DL framework to effectively perform feature selection on its own
via the training process, de-emphasizing unimportant descriptors; (2) as
discussed above, training time is small compared to dataset generation
time, or even compared to to the time it takes to compute the
descriptors using \texttt{mordred}.

\hypertarget{coverage-of-descriptors}{%
\subsection{Coverage of Descriptors}\label{coverage-of-descriptors}}

\texttt{fastprop} is fundamentally limited by the types of chemicals
which can be uniquely described by the \texttt{mordred} package.
Domain-specific additions which are not just derived from the
descriptors already implemented will be required to expand its
application to new domains.

For example, in its current state \texttt{mordred} does not include any
connectivity based-descriptors that reflect the presence or absence of
stereocenters. While some of the 3D descriptors it implements could
implicitly reflect sterochemistry, more explicit descriptors like the
Stereo Signature Molecular Descriptor {[}42{]} may prove helpful in the
future if re-implemented in \texttt{mordred}.

\hypertarget{interpretability}{%
\subsection{Interpretability}\label{interpretability}}

Though not discussed here for the sake of length, \texttt{fastprop}
contains the functionality to perform feature importance studies on
trained models. By using SHAP values {[}28{]} once can assign a scalar
‘importance’ to each of the input features with respect to the target
value, such as molecular weight having a significant positive impact on
boiling point in alkanes. Experts users can leverage this information to
guide molecular design and optimization or inform future lines of
inquiry. Via the \texttt{fastprop} CLI users can train a model and then
use \texttt{fastprop\ shap} to analyze the resulting trained network.
\texttt{fastprop\ shap} will then generate diagrams to visualize the
SHAP values.

\hypertarget{availability}{%
\section{Availability}\label{availability}}

\begin{itemize}
\tightlist
\item
  Project name: fastprop
\item
  Project home page: github.com/jacksonburns/fastprop
\item
  Operating system(s): Platform independent
\item
  Programming language: Python
\item
  Other requirements: pyyaml, lightning, mordredcommunity, astartes
\item
  License: MIT
\end{itemize}

\hypertarget{declarations}{%
\section{Declarations}\label{declarations}}

\hypertarget{availability-of-data-and-materials}{%
\subsection{Availability of data and
materials}\label{availability-of-data-and-materials}}

\texttt{fastprop} is Free and Open Source Software; anyone may view,
modify, and execute it according to the terms of the MIT license. See
github.com/jacksonburns/fastprop for more information.

All data used in the Benchmarks shown above is publicly available under
a permissive license. See the benchmarks directory at the
\texttt{fastprop} GitHub page for instructions on retrieving each
dataset and preparing it for use with \texttt{fastprop}, where
applicable.

\hypertarget{competing-interests}{%
\subsection{Competing interests}\label{competing-interests}}

None.

\hypertarget{funding}{%
\subsection{Funding}\label{funding}}

This material is based upon work supported by the U.S. Department of
Energy, Office of Science, Office of Advanced Scientific Computing
Research, Department of Energy Computational Science Graduate Fellowship
under Award Number DE-SC0023112.

\hypertarget{authors-contributions}{%
\subsection{Authors’ contributions}\label{authors-contributions}}

Initial ideation of \texttt{fastprop} was a joint effort of Burns and
Green. Implementation, benchmarking, and writing were done by Burns.

\hypertarget{acknowledgements}{%
\subsection{Acknowledgements}\label{acknowledgements}}

The authors acknowledge Haoyang Wu, Hao-Wei Pang, and Xiaorui Dong for
their insightful conversations when initially forming the central ideas
of \texttt{fastprop}.

\hypertarget{disclaimer}{%
\subsection{Disclaimer}\label{disclaimer}}

This report was prepared as an account of work sponsored by an agency of
the United States Government. Neither the United States Government nor
any agency thereof, nor any of their employees, makes any warranty,
express or implied, or assumes any legal liability or responsibility for
the accuracy, completeness, or usefulness of any information, apparatus,
product, or process disclosed, or represents that its use would not
infringe privately owned rights. Reference herein to any specific
commercial product, process, or service by trade name, trademark,
manufacturer, or otherwise does not necessarily constitute or imply its
endorsement, recommendation, or favoring by the United States Government
or any agency thereof. The views and opinions of authors expressed
herein do not necessarily state or reflect those of the United States
Government or any agency thereof.

\hypertarget{cited-works}{%
\section*{Cited Works}\label{cited-works}}
\addcontentsline{toc}{section}{Cited Works}

\hypertarget{refs}{}
\begin{CSLReferences}{0}{0}
\leavevmode\vadjust pre{\hypertarget{ref-muratov_qsar}{}}%
\CSLLeftMargin{1. }%
\CSLRightInline{Muratov EN, Bajorath J, Sheridan RP, et al (2020) QSAR
without borders. Chem Soc Rev 49:3525–3564.
\url{https://doi.org/10.1039/D0CS00098A}}

\leavevmode\vadjust pre{\hypertarget{ref-wiener_index}{}}%
\CSLLeftMargin{2. }%
\CSLRightInline{Wiener H (1947) Structural determination of paraffin
boiling points. Journal of the American Chemical Society 69:17–20.
\url{https://doi.org/10.1021/ja01193a005}}

\leavevmode\vadjust pre{\hypertarget{ref-estrada_abc}{}}%
\CSLLeftMargin{3. }%
\CSLRightInline{Estrada’ E, Torres’ L, Rodriguez’ L (1998)
\href{http://nopr.niscpr.res.in/handle/123456789/40308}{An atom-bond
connectivity index: Modelling the.enthalpy of formation of alkanes}.
Indian Journal of Chemistry 37:849–855}

\leavevmode\vadjust pre{\hypertarget{ref-descriptors_book}{}}%
\CSLLeftMargin{4. }%
\CSLRightInline{Todeschini R, Consonni V (2009)
\href{https://doi.org/10.1002/9783527628766}{Molecular descriptors for
chemoinformatics}. In: Methods and principles in medicinal chemistry. p
1252}

\leavevmode\vadjust pre{\hypertarget{ref-pah}{}}%
\CSLLeftMargin{5. }%
\CSLRightInline{Arockiaraj M, Paul D, Clement J, et al (2023) Novel
molecular hybrid geometric-harmonic-zagreb degree based descriptors and
their efficacy in QSPR studies of polycyclic aromatic hydrocarbons. SAR
and QSAR in Environmental Research 34:569–589.
\url{https://doi.org/10.1080/1062936x.2023.2239149}}

\leavevmode\vadjust pre{\hypertarget{ref-ma_deep_qsar}{}}%
\CSLLeftMargin{6. }%
\CSLRightInline{Ma J, Sheridan RP, Liaw A, et al (2015) Deep neural nets
as a method for quantitative structure-activity relationships. Journal
of Chemical Information and Modeling 55:263–274.
\url{https://doi.org/10.1021/ci500747n}}

\leavevmode\vadjust pre{\hypertarget{ref-Coley2017}{}}%
\CSLLeftMargin{7. }%
\CSLRightInline{Coley CW, Barzilay R, Jaakkola TS, et al (2017)
Prediction of organic reaction outcomes using machine learning. ACS
Central Science 3:434–443.
\url{https://doi.org/10.1021/acscentsci.7b00064}}

\leavevmode\vadjust pre{\hypertarget{ref-chemprop_theory}{}}%
\CSLLeftMargin{8. }%
\CSLRightInline{Yang K, Swanson K, Jin W, et al (2019) Analyzing learned
molecular representations for property prediction. Journal of Chemical
Information and Modeling 59:3370–3388.
\url{https://doi.org/10.1021/acs.jcim.9b00237}}

\leavevmode\vadjust pre{\hypertarget{ref-chemprop_software}{}}%
\CSLLeftMargin{9. }%
\CSLRightInline{Heid E, Greenman KP, Chung Y, et al (2024) Chemprop: A
machine learning package for chemical property prediction. Journal of
Chemical Information and Modeling 64:9–17.
\url{https://doi.org/10.1021/acs.jcim.3c01250}}

\leavevmode\vadjust pre{\hypertarget{ref-cmpnn}{}}%
\CSLLeftMargin{10. }%
\CSLRightInline{Song Y, Zheng S, Niu Z, et al (2021)
\href{https://doi.org/10.5555/3491440.3491832}{Communicative
representation learning on attributed molecular graphs}. In: Proceedings
of the twenty-ninth international joint conference on artificial
intelligence. Yokohama, Yokohama, Japan}

\leavevmode\vadjust pre{\hypertarget{ref-unimol}{}}%
\CSLLeftMargin{11. }%
\CSLRightInline{Zhou G, Gao Z, Ding Q, et al (2023)
\href{https://doi.org/10.26434/chemrxiv-2022-jjm0j}{Uni-mol: A universal
3D molecular representation learning framework}. In: The eleventh
international conference on learning representations}

\leavevmode\vadjust pre{\hypertarget{ref-mhnn}{}}%
\CSLLeftMargin{12. }%
\CSLRightInline{Chen J, Schwaller P (2023) Molecular hypergraph neural
networks. \url{https://doi.org/10.48550/arXiv.2312.13136}}

\leavevmode\vadjust pre{\hypertarget{ref-gslmpp}{}}%
\CSLLeftMargin{13. }%
\CSLRightInline{Zhao B, Xu W, Guan J, Zhou S (2023) Molecular property
prediction based on graph structure learning. arXiv preprint
arXiv:231216855}

\leavevmode\vadjust pre{\hypertarget{ref-sggrl}{}}%
\CSLLeftMargin{14. }%
\CSLRightInline{Wang Z, Jiang T, Wang J, Xuan Q (2024) Multi-modal
representation learning for molecular property prediction: Sequence,
graph, geometry. \url{https://doi.org/10.48550/ARXIV.2401.03369}}

\leavevmode\vadjust pre{\hypertarget{ref-moco}{}}%
\CSLLeftMargin{15. }%
\CSLRightInline{Zhu Y, Chen D, Du Y, et al (2024) Molecular contrastive
pretraining with collaborative featurizations. Journal of Chemical
Information and Modeling 64:1112–1122.
\url{https://doi.org/10.1021/acs.jcim.3c01468}}

\leavevmode\vadjust pre{\hypertarget{ref-deepdelta}{}}%
\CSLLeftMargin{16. }%
\CSLRightInline{Schaduangrat N, Anuwongcharoen N, Charoenkwan P,
Shoombuatong W (2023) DeepAR: A novel deep learning-based hybrid
framework for the interpretable prediction of androgen receptor
antagonists. Journal of Cheminformatics 15:50.
\url{https://doi.org/10.1186/s13321-023-00721-z}}

\leavevmode\vadjust pre{\hypertarget{ref-tcnn}{}}%
\CSLLeftMargin{17. }%
\CSLRightInline{Karpov P, Godin G, Tetko IV (2019) Transformer-CNN: Fast
and reliable tool for QSAR.
\url{https://doi.org/10.48550/ARXIV.1911.06603}}

\leavevmode\vadjust pre{\hypertarget{ref-low_data_review}{}}%
\CSLLeftMargin{18. }%
\CSLRightInline{Tilborg D van, Brinkmann H, Criscuolo E, et al (2024)
Deep learning for low-data drug discovery: Hurdles and opportunities.
ChemRxiv. \url{https://doi.org/10.26434/chemrxiv-2024-w0wvl}}

\leavevmode\vadjust pre{\hypertarget{ref-mordred}{}}%
\CSLLeftMargin{19. }%
\CSLRightInline{Moriwaki H, Tian Y-S, Kawashita N, Takagi T (2018)
Mordred: A molecular descriptor calculator. Journal of Cheminformatics
10:4. \url{https://doi.org/10.1186/s13321-018-0258-y}}

\leavevmode\vadjust pre{\hypertarget{ref-lightning}{}}%
\CSLLeftMargin{20. }%
\CSLRightInline{Falcon W, The PyTorch Lightning team (2019)
\href{https://doi.org/10.5281/zenodo.3828935}{{PyTorch Lightning}}}

\leavevmode\vadjust pre{\hypertarget{ref-smiles}{}}%
\CSLLeftMargin{21. }%
\CSLRightInline{Weininger D (1988) SMILES, a chemical language and
information system. 1. Introduction to methodology and encoding rules.
Journal of Chemical Information and Computer Sciences 28:31–36.
\url{https://doi.org/10.1021/ci00057a005}}

\leavevmode\vadjust pre{\hypertarget{ref-tetko_multitask}{}}%
\CSLLeftMargin{22. }%
\CSLRightInline{Sosnin S, Karlov D, Tetko IV, Fedorov MV (2019)
Comparative study of multitask toxicity modeling on a broad chemical
space. Journal of Chemical Information and Modeling 59:1062–1072.
\url{https://doi.org/10.1021/acs.jcim.8b00685}}

\leavevmode\vadjust pre{\hypertarget{ref-representation_review}{}}%
\CSLLeftMargin{23. }%
\CSLRightInline{McGibbon M, Shave S, Dong J, et al (2023) {From
intuition to AI: evolution of small molecule representations in drug
discovery}. Briefings in Bioinformatics 25:bbad422.
\url{https://doi.org/10.1093/bib/bbad422}}

\leavevmode\vadjust pre{\hypertarget{ref-fuels_qsar_method}{}}%
\CSLLeftMargin{24. }%
\CSLRightInline{Comesana AE, Huntington TT, Scown CD, et al (2022) A
systematic method for selecting molecular descriptors as features when
training models for predicting physiochemical properties. Fuel
321:123836.
https://doi.org/\url{https://doi.org/10.1016/j.fuel.2022.123836}}

\leavevmode\vadjust pre{\hypertarget{ref-wu_photovoltaic}{}}%
\CSLLeftMargin{25. }%
\CSLRightInline{Wu J, Wang S, Zhou L, et al (2020) Deep-learning
architecture in QSPR modeling for the prediction of energy conversion
efficiency of solar cells. Industrial and Engineering Chemistry Research
59:18991–19000. \url{https://doi.org/10.1021/acs.iecr.0c03880}}

\leavevmode\vadjust pre{\hypertarget{ref-fubrain}{}}%
\CSLLeftMargin{26. }%
\CSLRightInline{Esaki T, Ohashi R, Watanabe R, et al (2019)
Computational model to predict the fraction of unbound drug in the
brain. Journal of Chemical Information and Modeling 59:3251–3261.
\url{https://doi.org/10.1021/acs.jcim.9b00180}}

\leavevmode\vadjust pre{\hypertarget{ref-yalamanchi}{}}%
\CSLLeftMargin{27. }%
\CSLRightInline{Yalamanchi KK, Kommalapati S, Pal P, et al (2023)
Uncertainty quantification of a deep learning fuel property prediction
model. Applications in Energy and Combustion Science 16:100211.
https://doi.org/\url{https://doi.org/10.1016/j.jaecs.2023.100211}}

\leavevmode\vadjust pre{\hypertarget{ref-shap}{}}%
\CSLLeftMargin{28. }%
\CSLRightInline{Lundberg S, Lee S-I (2017) A unified approach to
interpreting model predictions.
\url{https://doi.org/10.48550/ARXIV.1705.07874}}

\leavevmode\vadjust pre{\hypertarget{ref-moleculenet}{}}%
\CSLLeftMargin{29. }%
\CSLRightInline{Wu Z, Ramsundar B, Feinberg EN, et al (2018)
MoleculeNet: A benchmark for molecular machine learning.
\url{https://doi.org/10.48550/arXiv.1703.00564}}

\leavevmode\vadjust pre{\hypertarget{ref-qm8}{}}%
\CSLLeftMargin{30. }%
\CSLRightInline{Ramakrishnan R, Hartmann M, Tapavicza E, von Liliendfeld
O (2015) Electronic spectra from TDDFT and machine learning in chemical
space. The Journal of Chemical Physics 143:
\url{https://doi.org/10.1063/1.4928757}}

\leavevmode\vadjust pre{\hypertarget{ref-qm9}{}}%
\CSLLeftMargin{31. }%
\CSLRightInline{Ramakrishnan R, Dral P, Rupp M, von Liliendfeld O (2014)
Quantum chemistry structures and properties of 134 kilo molecules.
Scientific Data 1: \url{https://doi.org/10.1038/sdata.2014.22}}

\leavevmode\vadjust pre{\hypertarget{ref-astartes}{}}%
\CSLLeftMargin{32. }%
\CSLRightInline{Burns J, Spiekermann K, Bhattacharjee H, et al (2023)
Machine learning validation via rational dataset sampling with astartes.
Journal of Open Source Software 8:5996.
\url{https://doi.org/10.21105/joss.05996}}

\leavevmode\vadjust pre{\hypertarget{ref-pgp_best}{}}%
\CSLLeftMargin{33. }%
\CSLRightInline{Notwell JH, Wood MW (2023) ADMET property prediction
through combinations of molecular fingerprints.
\url{https://doi.org/10.48550/arXiv.2310.00174}}

\leavevmode\vadjust pre{\hypertarget{ref-ara}{}}%
\CSLLeftMargin{34. }%
\CSLRightInline{Schaduangrat N, Anuwongcharoen N, Charoenkwan P,
Shoombuatong W (2023) DeepAR: A novel deep learning-based hybrid
framework for the interpretable prediction of androgen receptor
antagonists. Journal of Cheminformatics 15:50.
\url{https://doi.org/10.1186/s13321-023-00721-z}}

\leavevmode\vadjust pre{\hypertarget{ref-flash}{}}%
\CSLLeftMargin{35. }%
\CSLRightInline{Saldana DA, Starck L, Mougin P, et al (2011) Flash point
and cetane number predictions for fuel compounds using quantitative
structure property relationship (QSPR) methods. Energy \& Fuels
25:3900–3908. \url{https://doi.org/10.1021/ef200795j}}

\leavevmode\vadjust pre{\hypertarget{ref-ysi}{}}%
\CSLLeftMargin{36. }%
\CSLRightInline{Das DD, St. John PC, McEnally CS, et al (2018) Measuring
and predicting sooting tendencies of oxygenates, alkanes, alkenes,
cycloalkanes, and aromatics on a unified scale. Combustion and Flame
190:349–364. \url{https://doi.org/10.1016/j.combustflame.2017.12.005}}

\leavevmode\vadjust pre{\hypertarget{ref-pgp}{}}%
\CSLLeftMargin{37. }%
\CSLRightInline{Broccatelli F, Carosati E, Neri A, et al (2011) A novel
approach for predicting p-glycoprotein (ABCB1) inhibition using
molecular interaction fields. Journal of Medicinal Chemistry
54:1740–1751. \url{https://doi.org/10.1021/jm101421d}}

\leavevmode\vadjust pre{\hypertarget{ref-tdc}{}}%
\CSLLeftMargin{38. }%
\CSLRightInline{Huang K, Fu T, Gao W, et al (2021) Therapeutics data
commons: Machine learning datasets and tasks for therapeutics. CoRR
abs/2102.09548: \url{https://doi.org/10.48550/arXiv.2102.09548}}

\leavevmode\vadjust pre{\hypertarget{ref-tetko_overfit}{}}%
\CSLLeftMargin{39. }%
\CSLRightInline{Tetko IV, Deursen R van, Godin G (2024) Be aware of
overfitting by hyperparameter optimization! Journal of Cheminformatics
16:139. \url{https://doi.org/10.1186/s13321-024-00934-w}}

\leavevmode\vadjust pre{\hypertarget{ref-quantumscents}{}}%
\CSLLeftMargin{40. }%
\CSLRightInline{Burns JW, Rogers DM (2023) QuantumScents:
Quantum-mechanical properties for 3.5k olfactory molecules. Journal of
Chemical Information and Modeling 63:7330–7337.
\url{https://doi.org/10.1021/acs.jcim.3c01338}}

\leavevmode\vadjust pre{\hypertarget{ref-szymanski}{}}%
\CSLLeftMargin{41. }%
\CSLRightInline{Szymański P, Kajdanowicz T (2017)
\href{https://proceedings.mlr.press/v74/szyma\%C5\%84ski17a.html}{A
network perspective on stratification of multi-label data}. In: Luís
Torgo PB, Moniz N (eds) Proceedings of the first international workshop
on learning with imbalanced domains: Theory and applications. PMLR, pp
22–35}

\leavevmode\vadjust pre{\hypertarget{ref-stereo_signature}{}}%
\CSLLeftMargin{42. }%
\CSLRightInline{Carbonell P, Carlsson L, Faulon J-L (2013) Stereo
signature molecular descriptor. Journal of Chemical Information and
Modeling 53:887–897. \url{https://doi.org/10.1021/ci300584r}}

\end{CSLReferences}

\end{document}